\documentclass{article}

\PassOptionsToPackage{numbers, compress}{natbib}

\usepackage[preprint]{neurips_2024}

\usepackage[colorlinks = true,
            urlcolor  = black,
            linkcolor=darkgray]{hyperref}
\usepackage[utf8]{inputenc} 
\usepackage[T1]{fontenc}    
\usepackage{url}            
\usepackage{booktabs}       
\usepackage{amsfonts}       
\usepackage{nicefrac}       
\usepackage{microtype}      
\usepackage{xcolor}         

\usepackage{wrapfig}
\usepackage{graphicx}
\usepackage{multirow}
\usepackage{enumitem}
\usepackage{subcaption}

\usepackage{amsmath}
\usepackage{amssymb}
\usepackage{mathtools}
\usepackage{amsthm}
\usepackage{cleveref}

\theoremstyle{plain}
\newtheorem{theorem}{Theorem}[section]

\theoremstyle{definition}

\theoremstyle{remark}

\usepackage{amsmath,amsfonts,bm}
\usepackage{amssymb}
\usepackage{bbm}

\usepackage{listings}
\usepackage{xcolor}
\usepackage{courier}

\definecolor{codegreen}{rgb}{0,0.6,0}
\definecolor{codegray}{rgb}{0.5,0.5,0.5}
\definecolor{codeblack}{rgb}{0.,0.,0.}
\definecolor{codepurple}{rgb}{0.58,0,0.82}
\definecolor{backcolour}{rgb}{0.95,0.95,0.92}

\lstdefinestyle{mystyle}{
    backgroundcolor=\color{backcolour},   
    commentstyle=\color{codegreen},
    keywordstyle=\color{codeblack},
    numberstyle=\tiny\color{codegray},
    stringstyle=\color{codepurple},
    basicstyle=\ttfamily\footnotesize,
    breakatwhitespace=false,         
    breaklines=true,                 
    captionpos=b,                    
    keepspaces=true,                 
    numbers=left,                    
    numbersep=5pt,                  
    showspaces=false,                
    showstringspaces=false,
    showtabs=false,                  
    tabsize=2,
    aboveskip=0pt,
    belowskip=-3pt
}

\usepackage{amsthm}

\def\1{\bm{1}}

\def\vx{{\bm{x}}}

\def\mK{{\bm{K}}}

\def\mO{{\bm{O}}}

\def\mQ{{\bm{Q}}}

\def\mV{{\bm{V}}}

\def\mX{{\bm{X}}}

\DeclareMathAlphabet{\mathsfit}{\encodingdefault}{\sfdefault}{m}{sl}
\SetMathAlphabet{\mathsfit}{bold}{\encodingdefault}{\sfdefault}{bx}{n}

\def\sH{{\mathbb{H}}}

\NewDocumentEnvironment{mywrapfigure}{O{}mO{\wrapoverhang}mO{0pt}}{%
  \setlength{\intextsep}{#5}
  \wrapfigure[#1]{#2}[#3]{#4}%
}{%
  \endwrapfigure%
}

\title{Reasoning in Large Language Models: A Geometric Perspective}

\author{%
  Romain Cosentino$^*$, 
  Sarath Shekkizhar\thanks{Equal Contribution}\\
  \href{https://www.tenyx.com/research}{Tenyx}\\
  \texttt{\{romain,sarath\}@tenyx.com} 
}

\begin{document}
\maketitle
\begin{abstract}
The advancement of large language models (LLMs) for real-world applications hinges critically on enhancing their reasoning capabilities.
In this work, we explore the reasoning abilities of large language models (LLMs) through their geometrical understanding. We establish a connection between the expressive power of LLMs and the density of their self-attention graphs. Our analysis demonstrates that the density of these graphs defines the intrinsic dimension of the inputs to the MLP blocks. We demonstrate through theoretical analysis and toy examples that a higher intrinsic dimension implies a greater expressive capacity of the LLM. We further provide empirical evidence linking this geometric framework to recent advancements in methods aimed at enhancing the reasoning capabilities of LLMs.
\end{abstract}

\section{Introduction}


Large language models~(LLMs), such as GPT-4 \cite{achiam2023gpt}, Llama 3 \cite{llama3modelcard}, have achieved impressive performance on a wide range of tasks. The search for better LLMs hinges critically on the reasoning performance of these models. However, it is unclear what aspects of the language models are essential for achieving this goal. Today the predominant approach, considered by the community, to advance reasoning involves (i) increased model size (where larger models have resulted in better reasoning capabilities) \cite{kaplan2020scaling,hoffmann2022training,hernandez2021scaling} and (ii) increased context length \cite{pfau2024let}, more tokens or text as input to the LLM, through chain of thought \cite{wei2022chain}, retrieval augmented generation \cite{gao2024retrievalaugmentedgenerationlargelanguage}, or prompting with examples \cite{agarwal2024many}.

While these approaches have been sufficient, they represent only part of the potential avenues for improvement. Moreover, longer inputs and bigger models correspond to increased computational cost and inference latency for real-world use cases.  In this work, we take a principled approach to understand and elucidate the properties of LLMs that allow for improved and better reasoning. 
Our study leverages the geometry of the transformer layer \cite{vaswani2017attention}, a key component in LLMs, with empirical evidence on simulated as well as \textit{Llama 3} family of models \cite{llama3modelcard} to justify our claims. 

In particular, we characterize key properties of the transformer layer that are correlated with its capacity or expressive power. 
We show that the (i) density of interaction between the tokens in the self-attention or multi-head attention~(MHA) module of a transformer exemplifies the complexity of function representation achievable by the multi-layer perceptron~(MLP) layer that follows it, and (ii)  increased model size and context length facilitates higher attention density and consequently better reasoning. 
Our analysis presents a path toward improving reasoning and advancing LLMs while deepening our understanding of the models and their behaviors. We note that our accompanying work \cite{balestriero2023characterizing}, presented an analysis as in this work where we showed the brittleness of toxicity guardrails obtained via RLHF through the lens of LLM geometry.


In this work, we are specifically interested in understanding how the geometry of the LLM is correlated with its reasoning capabilities. Besides, we are investigating how increasing the input sequence length as well as the number of attention heads affects the geometry of the LLM. In fact, it has been empirically demonstrated that these are critical factors increasing LLMs' reasoning capabilities. 

We will start in \autoref{sec:THEORY} with a brief detour and highlight some important Deep Neural Networks (DNNs) geometric notions (\autoref{sec:DNN_EP}): $(i)$ how they are partitioning their input space and $(ii)$ how such a partitioning is related to their approximation capabilities. In this section, we will also show that increasing the intrinsic dimension of the DNN input affects its partitioning.

After this necessary detour, we will extrapolate these notions to LLMs (\autoref{sec:LLM}). We will highlight how one can capture their expressive power by inspecting the intrinsic dimension of the self-attention block. Specifically by analyzing the graph density of each attention head. We show how this intrinsic dimension is related to context length and the number of heads.

In \autoref{sec:EXP} we introduce a series of experiments designed to analyze the correlation between these geometrical properties and the LLM reasoning capabilities. 
Our findings reveal that as the number of examples provided in the prompt increases, the intrinsic dimension of the LLM also rises. Notably, while the increase in the intrinsic dimension at the first layer is not indicative of the accuracy of the model's responses, a significant rise in the intrinsic dimension at the final layer strongly correlates with enhanced reasoning performance. This suggests that the geometry of the LLM’s internal representations plays a crucial role in its ability to reason effectively.

\section{Input Space Partitioning and Expressive Power}
\label{sec:THEORY}
In this section, we delve into the geometrical intuitions that underpin a fundamental aspect of Deep Neural Networks (DNNs): the adaptive partitioning of the DNN input space. This process leads to the formation of regions within the input space, each associated with an affine map that characterizes how the network processes the inputs in that region. We then leverage this perspective in conjunction with the multi-head attention~(MHA) layer in the transformer module to develop a novel geometric view of LLMs. This perspective allows us to hypothesize about the role of model size and context length in modern LLMs and presents a path toward alternate ideas that can lead to improved reasoning capabilities.

\subsection{Deep Neural Networks}
\label{sec:DNN_EP}
We describe the continuous piecewise affine formulation of DNNs to elucidate the concept of their induced local linear mappings. In particular, we focus on the simple case of the multilayer perceptron~(MLP) consisting of one hidden layer, typically employed in a transformer, from a spline geometric viewpoint. Subsequently, we provide an intuitive depiction through simulated experiments of their approximation capabilities, emphasizing the significance of the adaptive partitioning property, and the role of input space dimension.

\begin{figure}[th]
\centering
\begin{minipage}{.45\textwidth}
    \centering
    \includegraphics[width=1\linewidth]{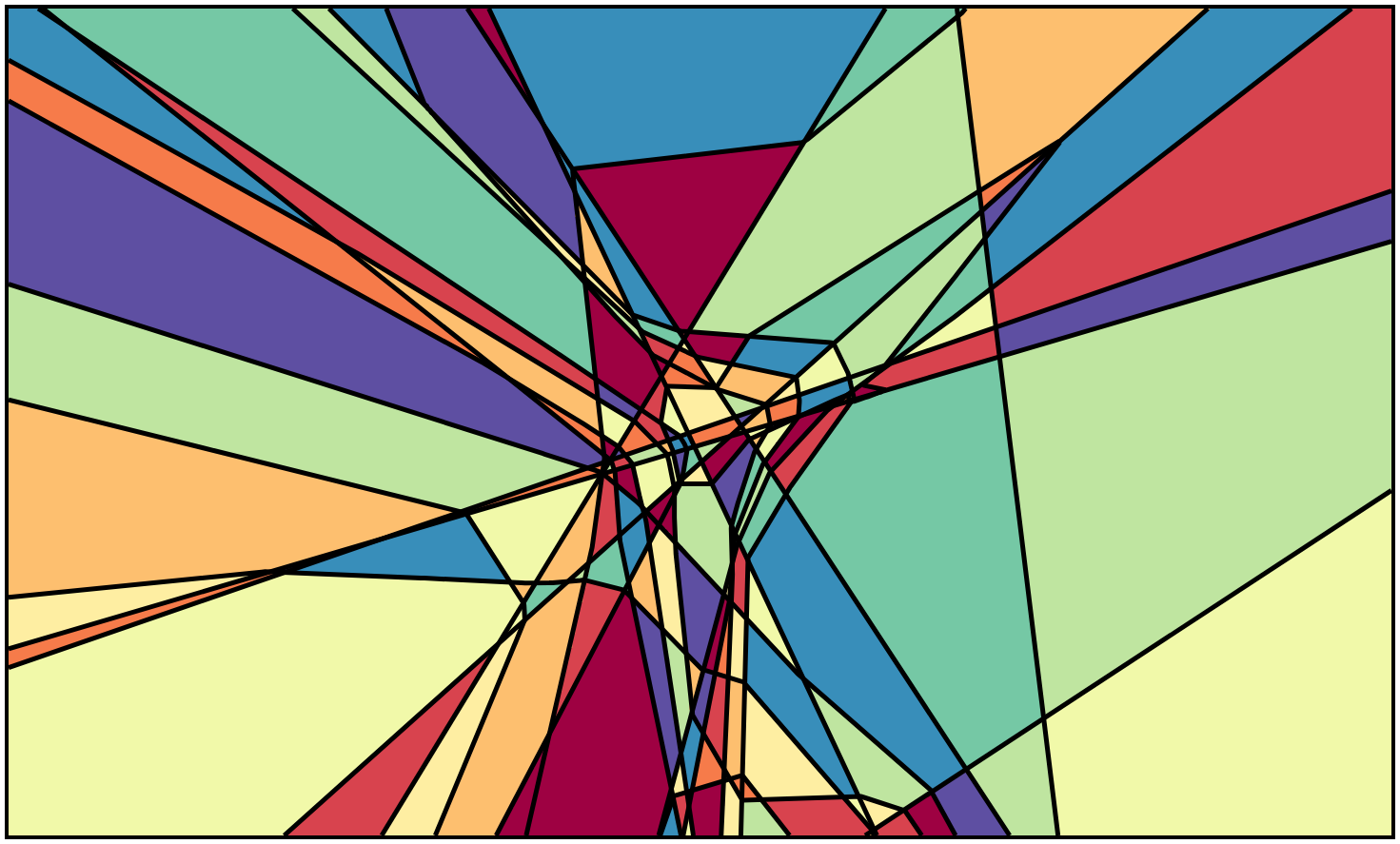}
\end{minipage}
\hspace{1cm}
\begin{minipage}{.45\textwidth}
    \centering
    \includegraphics[width=1\linewidth]{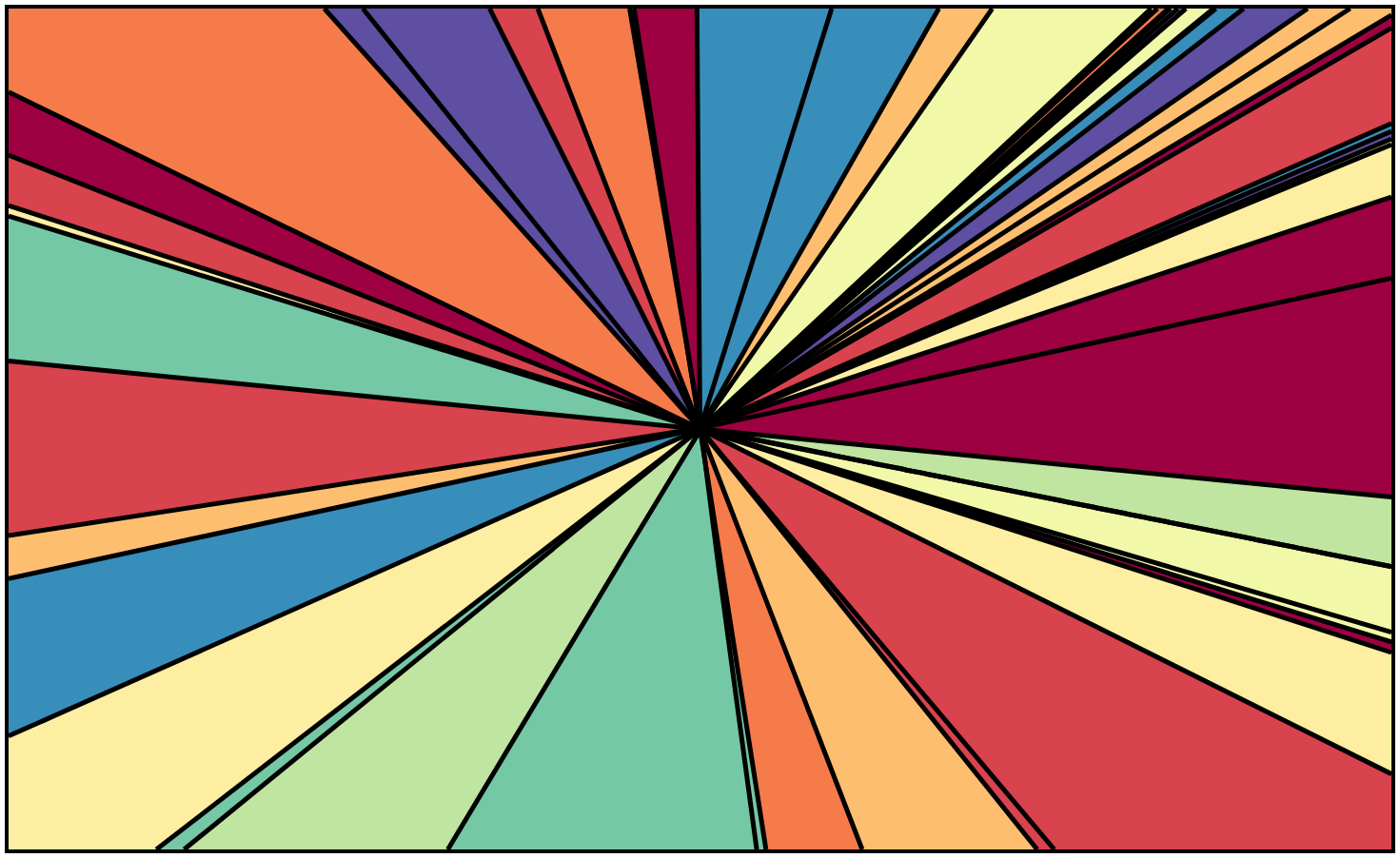}
\end{minipage}
    \caption{\textbf{Continuous Piece-wise Affine view of MLP}. 2-dimensional visualization of the input space partitioning induced by a one hidden layer MLP randomly initialized using standard with bias (\textbf{Left}) and zero bias (\textbf{Right}). 
    Each region, depicted by a particular color and bounded by black lines, has a set of CPA parameters $A_{\omega}, B_{\omega}$ described in \autoref{eq:cpa}. These parameters depend on the per-layer affine parameters and the state of the nonlinearities of the region $\omega$. }
    \label{fig:inputspacepartitioning}
\end{figure}

\paragraph{Continuous Piece-wise Affine Formulation of DNNs:}
The geometric characterization of MLPs employing nonlinearities, such as (leaky-)ReLU, absolute value, and max-pooling, have been extensively studied from the lens of a continuous piecewise linear operator, resulting in a partition $\Omega$ of the input space \cite{pmlr-v80-balestriero18b, balestriero2020mad,balestriero2019geometry}. As such, a DNN defined as  $\boldsymbol{f}_\Theta$ with parameters $\Theta$ can be re-written as
\begin{equation}
\label{eq:cpa}
    \boldsymbol{f}_{\Theta}(\boldsymbol{x}) = \sum_{\omega \in \Omega} 1_{\{\boldsymbol{x} \in \omega\}}\left(A_{\omega}\boldsymbol{x}+B_{\omega} \right),
\end{equation}
where $1$ defines the indicator function, $A_{\omega}$ and $B_{\omega}$ the per region affine parameters associated with the DNN layer, and $\boldsymbol{x}$ the input to the network. The indicator function is data dependent and subsumes the affine parameters and the nonlinearity of the region $\omega \in \Omega$. A depiction of the regions and the partition induced by an MLP having a $2$-dimensional input is given in \autoref{fig:inputspacepartitioning}.

\paragraph{Partitioning, Number of regions, and Function approximation:}
The approximation capability of a DNN for a given interval in the input space is directly proportional to the number of regions and the mapping associated with that input space interval. As per the continuous piece-wise affine property of DNNs defined in \autoref{eq:cpa}, consider the two possible scenarios in terms of approximation:  $(i)$  the target function is linear in a given interval, in which case a single region is sufficient enough to approximate it; the DNN is only required to adjust its slope and bias for the interval, or $(ii)$ the function is non-linear in the interval, in which case the DNN would require multiple regions to approximate the curvature of the target function; the more regions in the interval corresponds in turn to better function approximation.

\begin{figure}[h] 
  \begin{minipage}[b]{0.5\linewidth}
    \centering
        \small{\hspace{.35cm} \textit{Number hidden neurons: $50$}}
        \vspace{0.2cm}
    \includegraphics[width=1\linewidth]{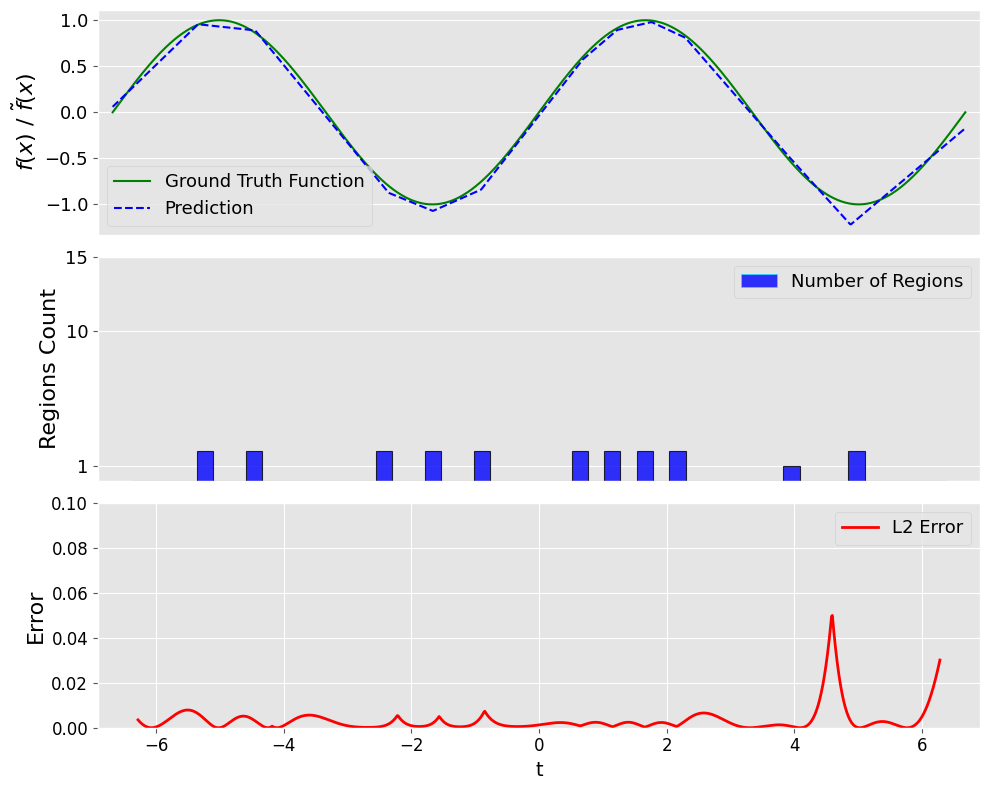}
  \end{minipage}
  \begin{minipage}[b]{0.5\linewidth}
    \centering
            \small{\hspace{.35cm} \textit{Number hidden neurons: $500$}}
        \vspace{0.2cm}
    \includegraphics[width=1\linewidth]{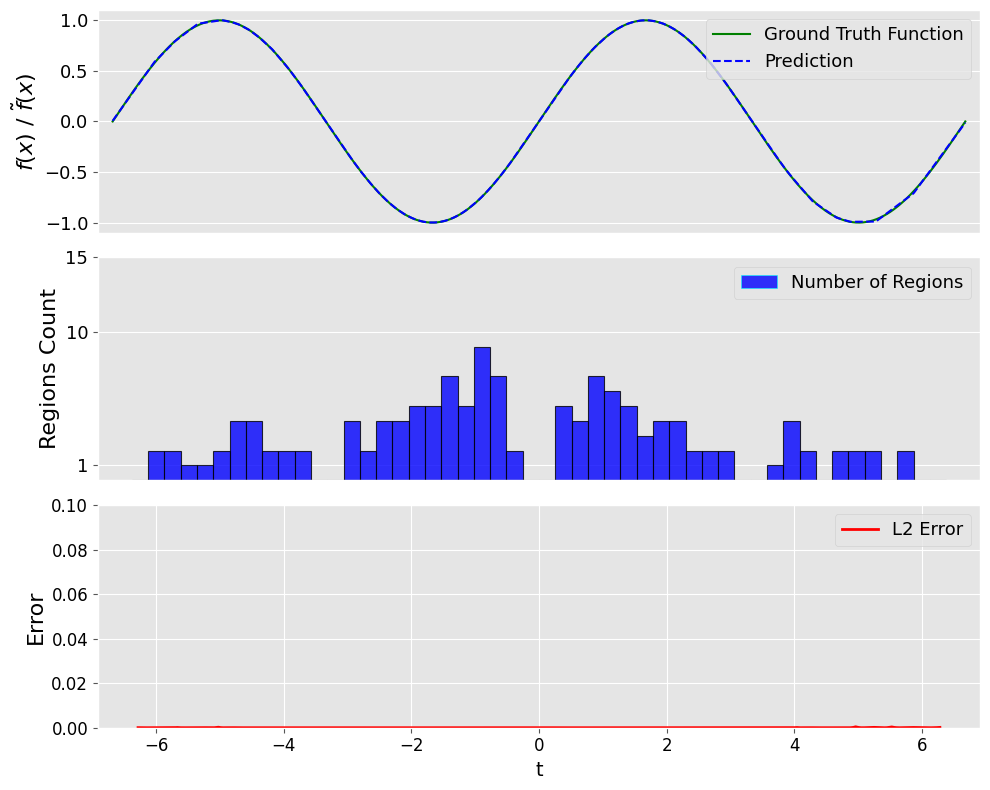}
  \end{minipage} 
  \caption{\textbf{DNN approximation \& induced number of input space regions}. The ground truth and approximation of a sine function by an MLP ( (\textbf{Top})), the number of associated regions the MLP induces in its input space (\textbf{Middle}), and the approximation error (\textbf{Bottom}). We present results for a $1$-hidden layer MLP with $50$ neurons (\textbf{Left}) and $500$ neurons (\textbf{Right}). We note that the model breaks from its linear behavior with the DNN introducing a new region whenever a change of direction in the MLP mapping occurs. Subsequently, we obtain a new affine mapping as per \autoref{eq:cpa} for each new region created by the model with finer approximation in spaces where the number of regions is higher, as seen in the wider MLP with $500$ neurons. The crucial advantage of DNNs is their ability to adapt the positioning of these regions and learn data-driven partitions. 
}
  \label{fig:sine_approx} 
\end{figure}

In \autoref{fig:sine_approx}, we validate the above claim and present a visualization of such phenomena in DNN. The target function to be approximated is a simple $\sin$ function with input space  $\in [-2\pi, 2\pi ]$. First, the higher the number of neurons, the higher the approximation power. In particular, with enough regions, the DNN can approximate arbitrarily complex functions within an input space. Theoretically, we know that a DNN with an infinite number of neurons is a universal approximator and the geometric view presents a different view of the same theorem. Second, the approximation error associated with each interval locally is directly proportional to the number of regions available to the DNN in that interval. Finally, the positioning of these regions is data-driven, albeit architectural changes induce a bias, DNNs can densify with more or less partitions in their input space that require more curvature based on the uniformity and size of training data. 

\begin{wrapfigure}{r}{0.5\textwidth} 
    \centering
    \includegraphics[width=1\linewidth]{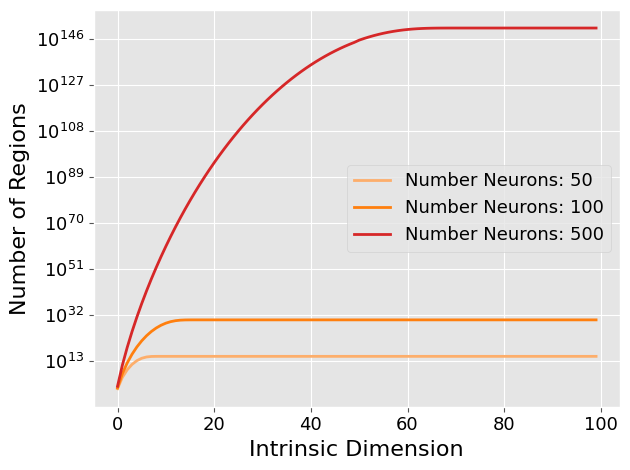}
   \caption{\textbf{Number of regions as a function of input dimension} - Upper bound of number of regions spanned by a $1$-hidden layer MLP ($50$, $100$, and $500$ neurons) concerning the input space intrinsic dimension. We observe that increasing the intrinsic dimension affect exponentially the number of regions. As such, for a given number of neurons, one can artificially increase the number of regions by increasing the intrinsic dimension of the input space. This will be a crucial component to understanding why increasing the size of the prompt via many-shot or CoT induces better reasoning capabilities in LLMs. This will be the central point of \autoref{sec:LLM} as well as \autoref{sec:EXP}. }
    \label{fig:num_regions} 
\end{wrapfigure}

When adding neurons there is an increase in the number of regions, thus the approximation power of the DNN does increase. We ask now the question of whether there is another way to increase DNN capacity without affecting the architecture. In particular, we investigate how the number of regions interacts with the \emph{intrinsic dimension}~(see \autoref{sec:LLM} for definition) of the input space. In \autoref{fig:num_regions}, we show for different sizes of $1$-hidden layer MLP that the number of regions scales exponentially with the intrinsic dimension. 


In the following section, we make use of the geometrical aspects of MLPs, i.e., the approximation, expressivity, and dimensionality, in conjunction with a multi-head attention layer to understand the geometry of transformer modules in LLMs. In particular, we present a framework for understanding LLMs through these geometrical concepts, both from a theoretical and empirical standpoint.

\subsection{Large Language Models}
\label{sec:LLM}
In this section, we interpret the architectural components of an LLM and its variations that can help improve the expressive power of the LLMs. Concretely, we will study the impact of the LLM-induced partition concerning an increase in the number of attention heads as well as the context length~(the sequence of tokens passed as input). To do so, we will exploit results from \cite{balestriero2023characterizing}, showing that the expressive power of an LLM increases as the intrinsic dimension of the self-attention layer increases. 

\paragraph{Intrinsic dimension $\propto$ Multi-Head Attention graph density:}
We begin by introducing notation through the definition of a transformer layer in a causal LLM, as follows 
\begin{align}
&\hspace{0.5em} {\rm Head}^{(\ell)}_h(\mX) \triangleq {\rm softmax_{causal}}\left(\mX\mQ_{h}^{(\ell)}\left(\mX\mK_{h}^{(\ell)}\right)^\top\right)\mX\mV_{h}^{(\ell)},&& \hspace{-.5cm} \text{(single-head mapping of $\mX$)}\label{eq:head}\\
&\hspace{0.5em} {\rm MHA}^{(\ell)}(\mX) \triangleq \sum_{h=1}^{H}{\rm Head}^{(\ell)}_h(\mX)\mO_h^{(\ell)},&& \hspace{-.2cm} \text{(combination of $H$ heads)}\label{eq:multihead}\\
&\hspace{0.5em} {\rm Layer}^{(\ell)}(\mX) \triangleq {\rm MLP}^{(\ell)}\left({\rm LayerNorm}^{(\ell)}\left({\rm MHA}^{(\ell)}(\mX)+\mX\right)\right)+\mX, &&\text{(single layer)}\label{eq:block}\\
&\hspace{0.5em} {\rm LLM}(\mX) \triangleq \left( {\rm Layer}^{(L)} \circ \dots \circ {\rm Layer}^{(1)} \right) (\mX),&&\text{(compose L layers)}\label{eq:LLM}
\end{align}
where we denote the attention map as follows
\begin{align}
    {\rm Attn}_h^{(\ell)}(\mX) \triangleq {\rm softmax_{causal}}\left(\mX\mQ_{h}^{(\ell)}{\mK_{h}^{(\ell)}}^\top\mX^\top\right)\label{eq:attn}.
\end{align}

It is evident from \autoref{eq:attn}, that the output of an attention layer is a right stochastic matrix that defines a graph where the nodes of the graph are the sequence of tokens and the edges (weights) are defined by the attention values. We will usually refer to \textit{density} of the self-attention graph when expressing the level of connectivity of the graph, i.e., the number of tokens that have an edge. 

In \autoref{prop:multihead}, we capture explicitly the relationship between the output of the multi-head attention layer as defined in \autoref{eq:LLM} and the intrinsic dimension driven by the sum of the dimensions induced in each individual attention layer. 

\begin{theorem}[causal multi-head Minkowski sum (\cite{balestriero2023characterizing}) ]
    \label{prop:multihead}
    The $i^{\rm th}$ row of the MHA mapping output (\autoref{eq:multihead}) lives in the Minkowski sum of single-head convex hulls  as
    $
({\rm MHA}^{(\ell)}(\mX)_{i,.})^\top \in \sH^{(\ell)}_1(i)  + \dots + \sH^{(\ell)}_H(i)
   $
   where $\sH^{(\ell)}_h(i) \triangleq {\rm Hull}\left\{(\mV_h^{(\ell)}\mO_h^{(\ell)})^\top \vx_j, j=1,\dots,i\right\}$
   with effective dimension at most
    \begin{align}
\sum_{h=1}^{H}\#\left\{{\rm Attn}_h^{(\ell)}(\mX^{(\ell)})_{i,j}>0,\; j=\{1, 2,\dots, i\}\right\}.\label{eq:dim}
    \end{align}
\end{theorem}

From \autoref{eq:dim}, it is clear that the intrinsic dimension can be increased by either \textit{$(i)$ enforcing a highly connected attention graph, or $(ii)$ adding more attention heads}. We will now exploit such a property and connect it to the expressive power of LLMs.

\paragraph{Intrinsic Dimension (ID):}The ID of an embedding space refers to the minimum number of parameters required for its characterization while maintaining its structure~\cite{bennett1969intrinsic}. 
Approaches for ID estimation~\cite{campadelli2015intrinsic,pope2021intrinsic} often rely on the construction of similarity-based graphs \cite{shekkizhar2020graph}. However, in LLMs, the similarity graph is readily available in the form of attention values. 
We define a soft notion of intrinsic dimension, equivalent to the definition in \autoref{prop:multihead}, namely,
\begin{align}
   \hspace{-0.15cm} \textbf{ID}^{\ell}_{\epsilon}(i) \hspace{-0.05cm}  \hspace{-0.05cm} \coloneqq  \hspace{-0.05cm} \hspace{-0.05cm}\#\left\{{\rm Attn}_h^{(\ell)}(\mX^{(\ell)})_{i,j}\hspace{-0.05cm} >\hspace{-0.05cm} \epsilon,\; \hspace{-0.05cm} j \hspace{-0.05cm}= \hspace{-0.05cm}\{1,2, \dots, i\} \right\}. \hspace{-0.05cm}
   \label{eq:id_est}
\end{align}
Intuitively, $\textbf{ID}^{\ell}_{\epsilon}(i)$ is the number of tokens that are influential, beyond a threshold  $\epsilon$, in defining the $i^{th}$ embedding. In practice, we set the threshold $\epsilon$ based on the statistics and the distribution of the attention values across several examples ($0.1$ in all our experiments).

\begin{figure}[ht] 
  \begin{minipage}[b]{0.5\linewidth}
    \centering
        \small{\hspace{.35cm} \textit{Context Length: $10$ - Numbers of Heads: $1$}}
        \vspace{0.2cm}
    \includegraphics[width=1\linewidth]{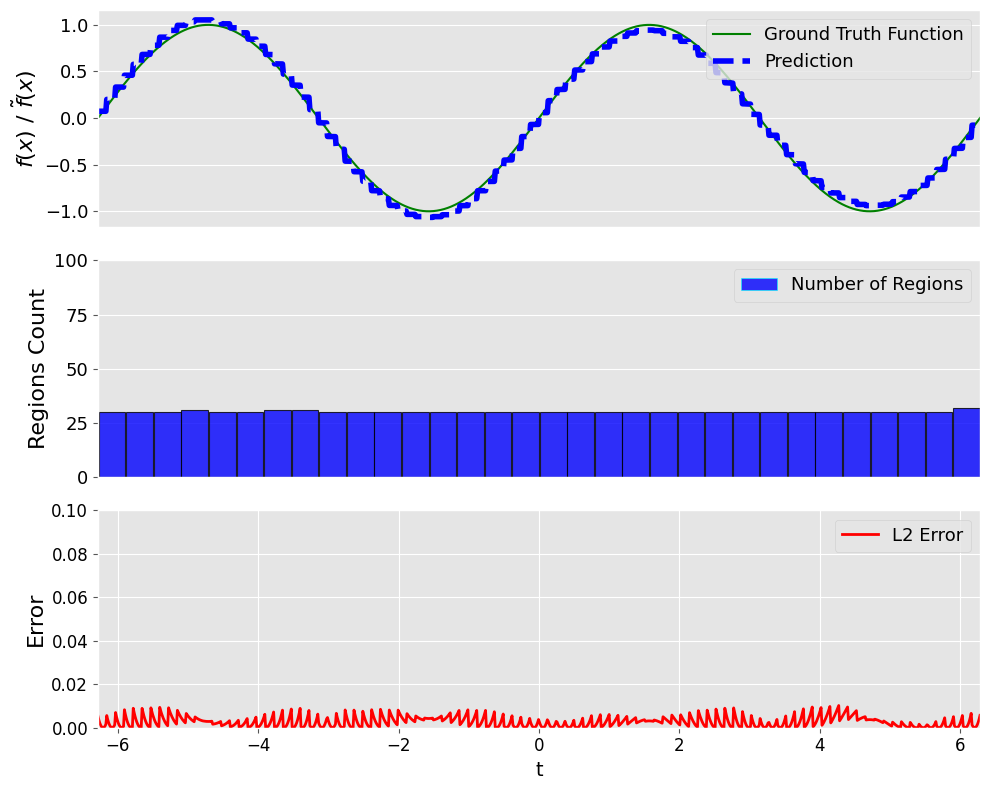}
  \end{minipage}
  \begin{minipage}[b]{0.5\linewidth}
    \centering
            \small{\hspace{.35cm} \textit{Context Length: $10$ - Numbers of Heads: $10$}}
        \vspace{0.2cm}
    \includegraphics[width=1\linewidth]{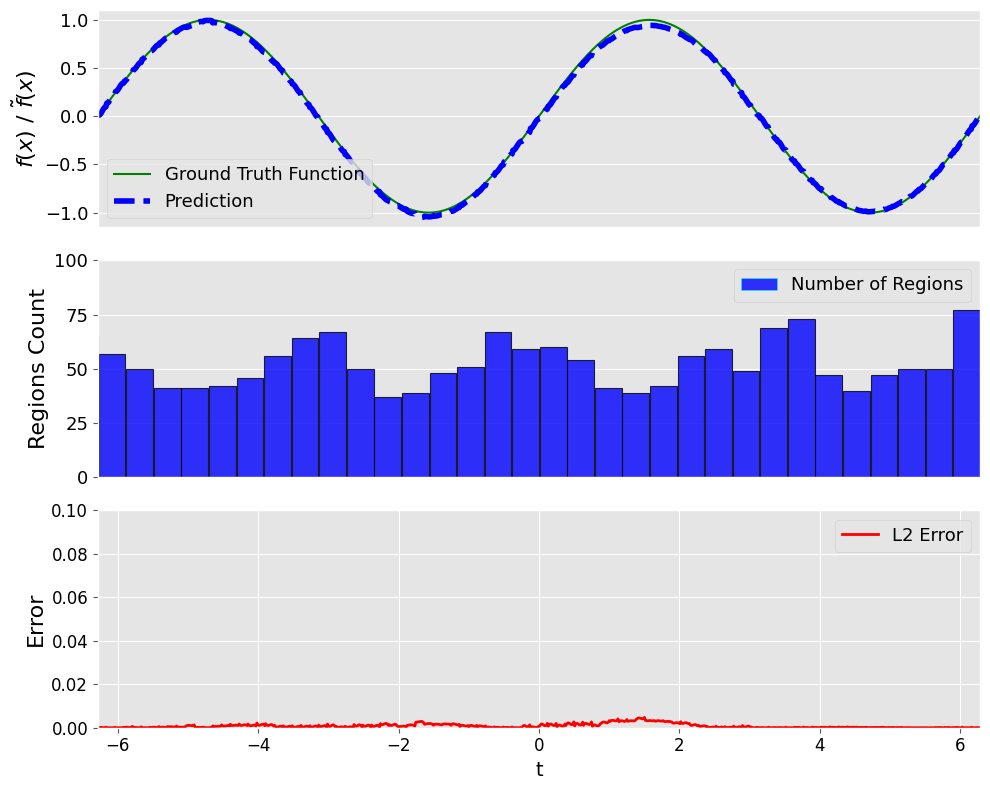}
  \end{minipage} 
    \begin{minipage}[b]{0.5\linewidth}
    \centering
                \small{\hspace{.35cm} \textit{Context Length: $100$ - Numbers of Heads: $1$}}
        \vspace{0.2cm}
    \includegraphics[width=1\linewidth]{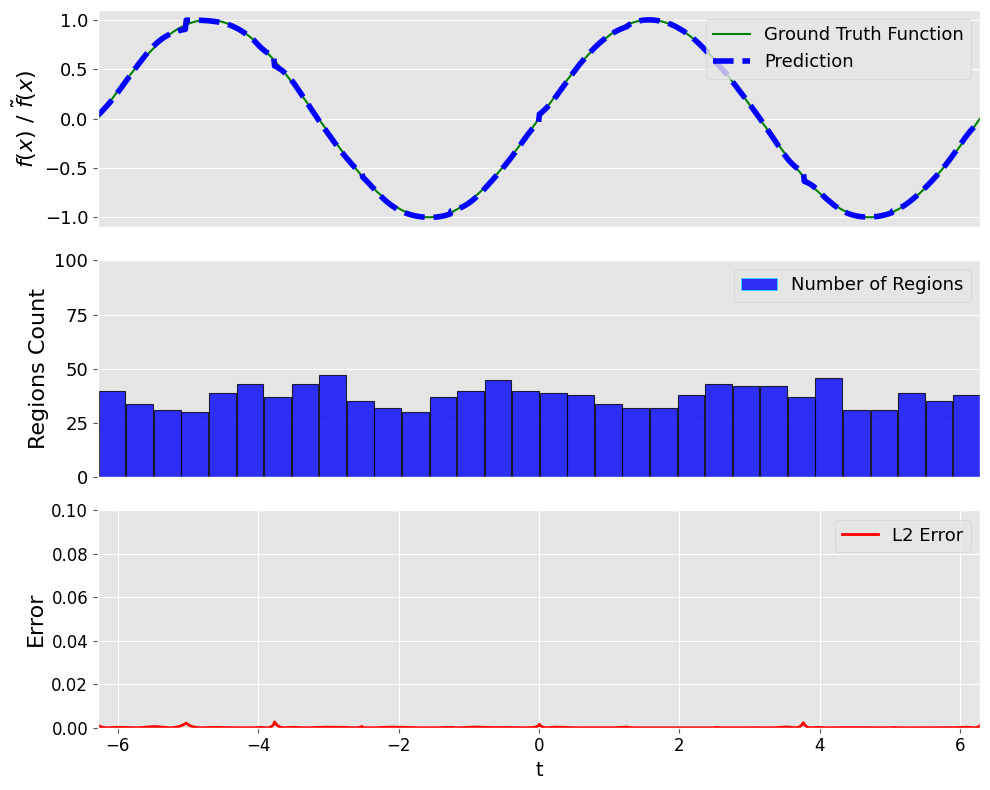}
  \end{minipage}
  \begin{minipage}[b]{0.5\linewidth}
    \centering
                \small{\hspace{.35cm} \textit{Context Length: $100$ - Numbers of Heads: $10$}}
        \vspace{0.2cm}
    \includegraphics[width=1\linewidth]{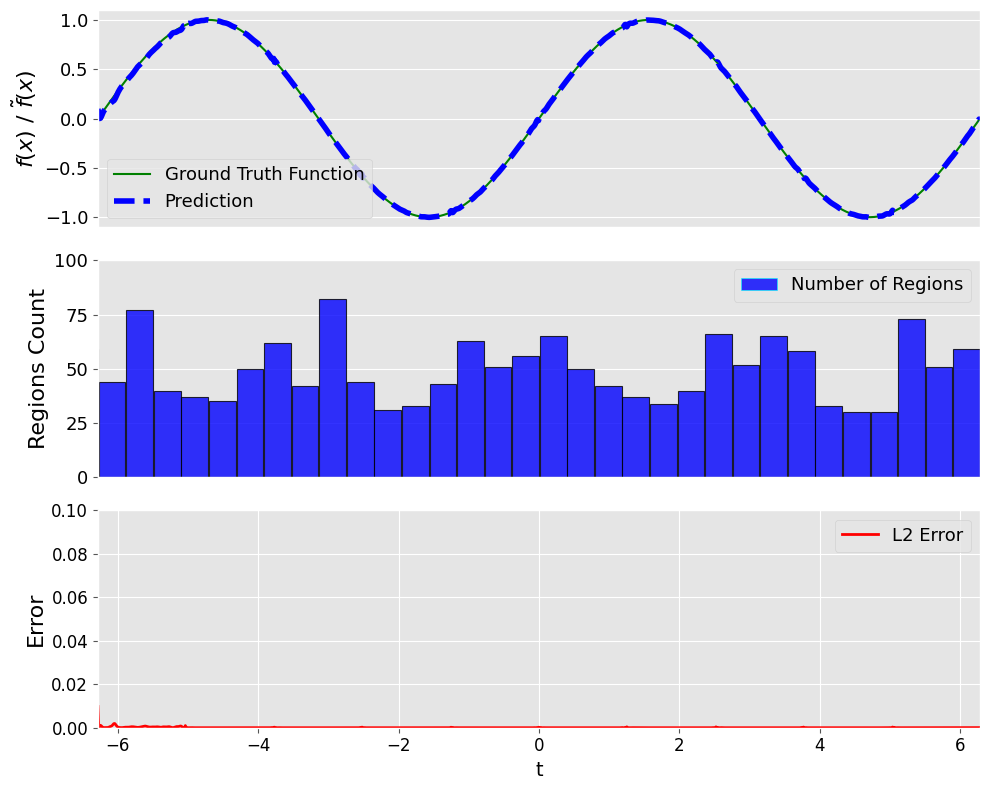}
  \end{minipage} 
  \caption{\textbf{LLM approximation \& induced number of input space regions} - Visualization of $sin(t)$ ($1000$ time bins) approximation by a $1$-block LLM, i.e., embedding $\rightarrow$ attention block (as in \autoref{eq:multihead}) $\rightarrow$ $1$-hidden layer MLP. We display the approximation of the sin function together with the number of regions induced by the MLP in the input space for different numbers of heads and context lengths \textit{(Top Left}) context length: $10$ and number of heads: $1$, \textit{(Top Right}) context length: $10$ and number of heads: $10$, \textit{(Bottom Left}) context length: $100$ and number of heads: $1$, \textit{(Bottom Right}) context length: $100$ and number of heads: $10$. We observe that both context length and number of heads are inducing an increase in the number of regions spanned by the MLP in the input space, which improves the approximation capabilities of the LLM. This result coincides with our geometrical description.
}
\label{fig:llm_sine_approx} 
\end{figure}

\paragraph{LLM expressive power $\propto$ intrinsic dimension:}
\autoref{prop:multihead} is consequential, specifically when we consider \autoref{sec:DNN_EP}, and in particular with \autoref{fig:num_regions}. We showed that: $(i)$ the higher the number of regions, the higher the approximation capability of DNNs, and $(ii)$ the number of regions can be increased by, not only having more neurons but by \textit{increasing the ID of the MLP's input}. 

We also know from the transformer architecture described in \autoref{eq:head} through \autoref{eq:LLM} and \autoref{prop:multihead} that the intrinsic dimension of the input to the MLP is driven by the attention maps. Therefore, the higher the density of the attention graph, the higher the number of regions that will be induced by the MLP, and thus, the higher its expressive power. 

\begin{figure}[ht] 
    \centering
    \begin{minipage}{.48\textwidth}
    \includegraphics[width=1\linewidth]{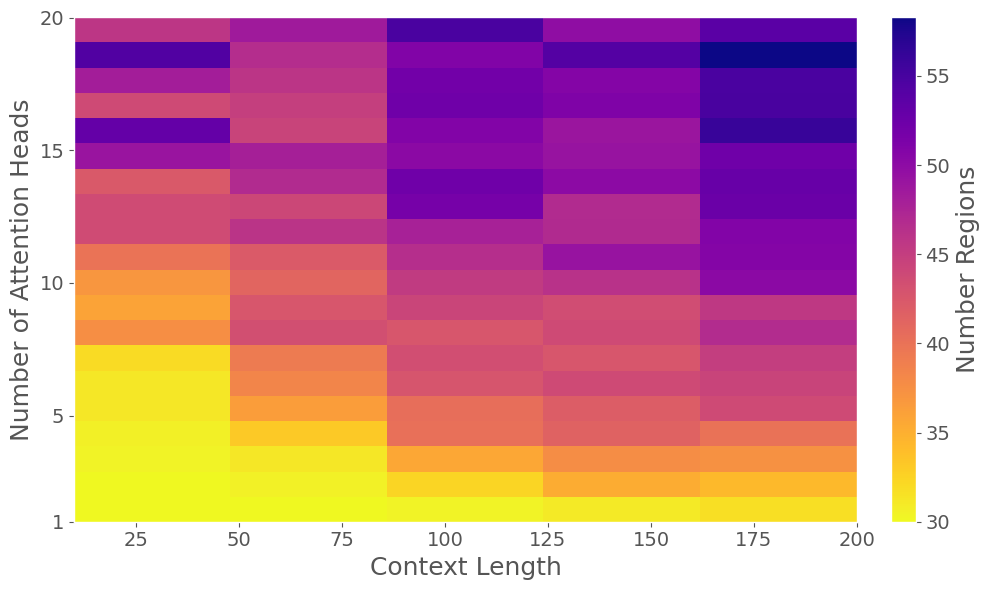}
    \end{minipage}
        \begin{minipage}{.48\textwidth}
    \includegraphics[width=1\linewidth]{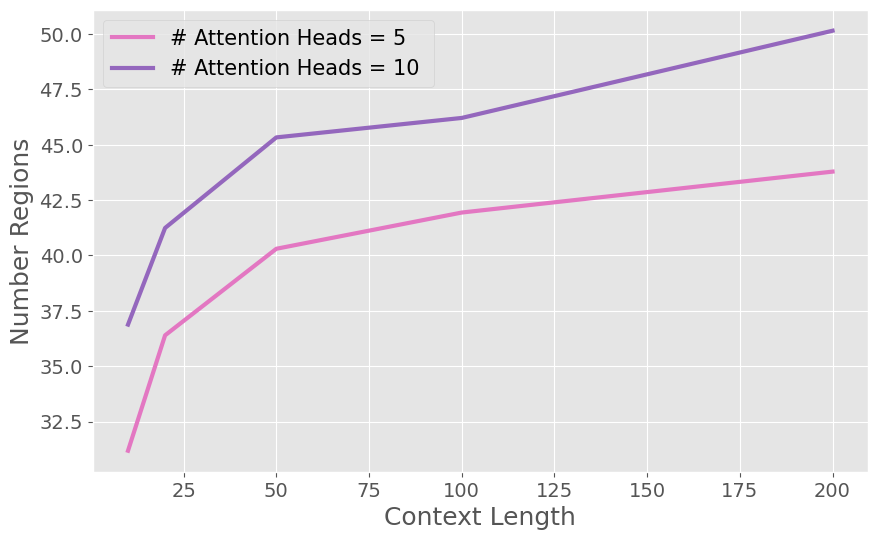}
    \end{minipage}
  \caption{\textbf{LLM input space regions} - (\textbf{Left})  Depiction of the number of regions induced by the MLP block in the input space of the LLM concerning the number of attention heads and context length. (\textbf{Right}) Zoom in on two rows of the left figure, specifically for several attention heads: $5,10$.  We observe that increasing both attention heads and context length does increase the number of regions, which as mentioned, leads to better approximation properties. It is important to note that, while changing the number of attention heads can be tedious and require pre-training or fine-tuning,  one can seamlessly vary the context length. There is therefore a way to improve LLM approximation capability without interacting with the weights of the model. }
\label{fig:region_quant} 
\end{figure}

It is now clear that one can increase the expressive power of an LLM by $(i)$ increasing the number of heads as per the additive nature of  \autoref{eq:dim}, $(ii)$ performing prompt modifications as to increase the density of the attention graph. Note that both these approaches have commonly been employed in various aspects in the last couple of years.

In \autoref{fig:llm_sine_approx}, we propose to re-use our sine function toy-example. Specifically, we show the number of regions induced by the MLP for different context lengths and number of heads. We consider a one-layer LLM, i.e., embedding, self-attention, and then $1$-hidden layer MLP. To encode the $1$-dimensional time dimension into a higher dimensional space, we consider the embedding layer a "positional encoding". Specifically, each time bin $t$ is mapped to a sinusoid which frequency depends on the context length as well as the position. We observe that the number of regions induced by the MLP in the input space increases with both the context length and the number of heads. Similarly as with the MLP example in \autoref{sec:DNN_EP}, the capabilities of the LLM are tied to the number of regions, that is, the more populated a region of the input space, the better the approximation.

We provide in \autoref{fig:region_quant} a more quantitative experiment regarding the number of regions induced by the MLP concerning context length and number of attention heads. Here again, we observe that to increase the number of regions and therefore improve the approximation capabilities of LLMs, one can increase the number of heads in the self-attention block or increase the context length.

It is now clear that these correlations are the result of \autoref{prop:multihead} together with the hyperplane arrangement result displayed in \autoref{fig:num_regions}. That is, the number of regions induced by hyperplane arrangement exponentially increases with high intrinsic dimensional spaces. In LLMs we identified that the number of heads as well as the context length are ways to increase the intrinsic dimension of the MLP input, therefore increasing its capability to generate dense partitions. 

We now propose to analyze how using this geometrical relationship as a tool to increase the expressive power of LLM can lead to better reasoning capabilities. 

\section{Experiment: Increasing LLM expressive power does improve its reasoning ability}
\label{sec:EXP}

\begin{figure}[hbp]

    \begin{subfigure}{0.48\textwidth}
        \centering
        \textit{Llama3 8B}
        \vspace{0.2cm}
        \includegraphics[width=\textwidth]{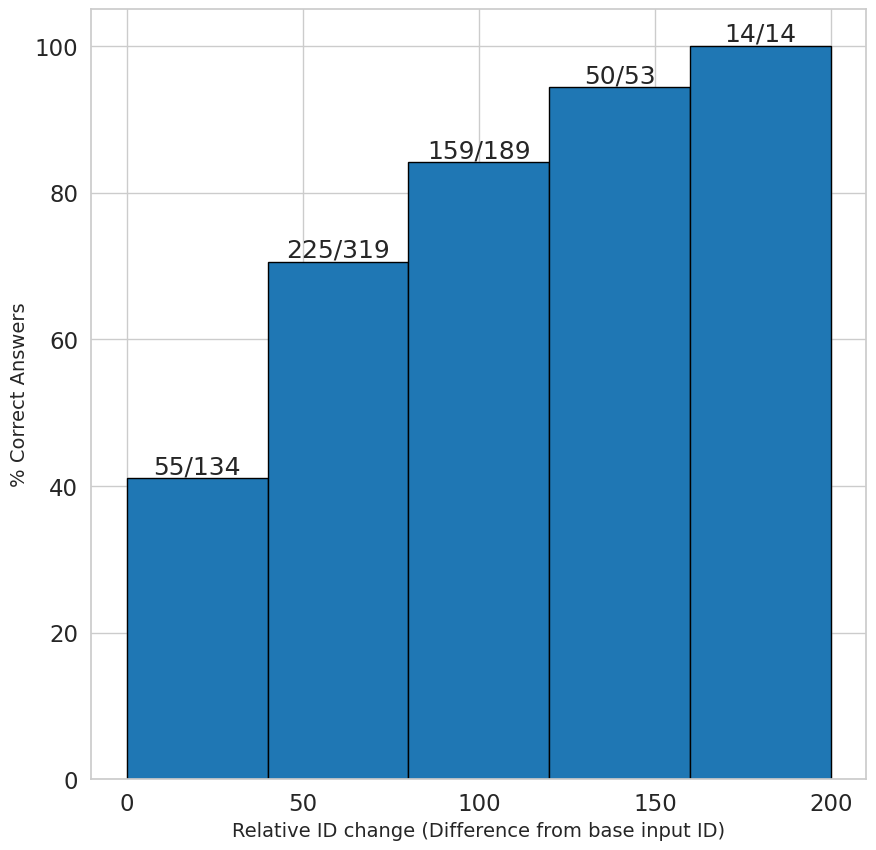}
    \end{subfigure}
    ~
    \begin{subfigure}{0.48\textwidth}
        \centering
        \textit{Llama3 70B}
        \vspace{0.2cm}
        \includegraphics[width=\textwidth]{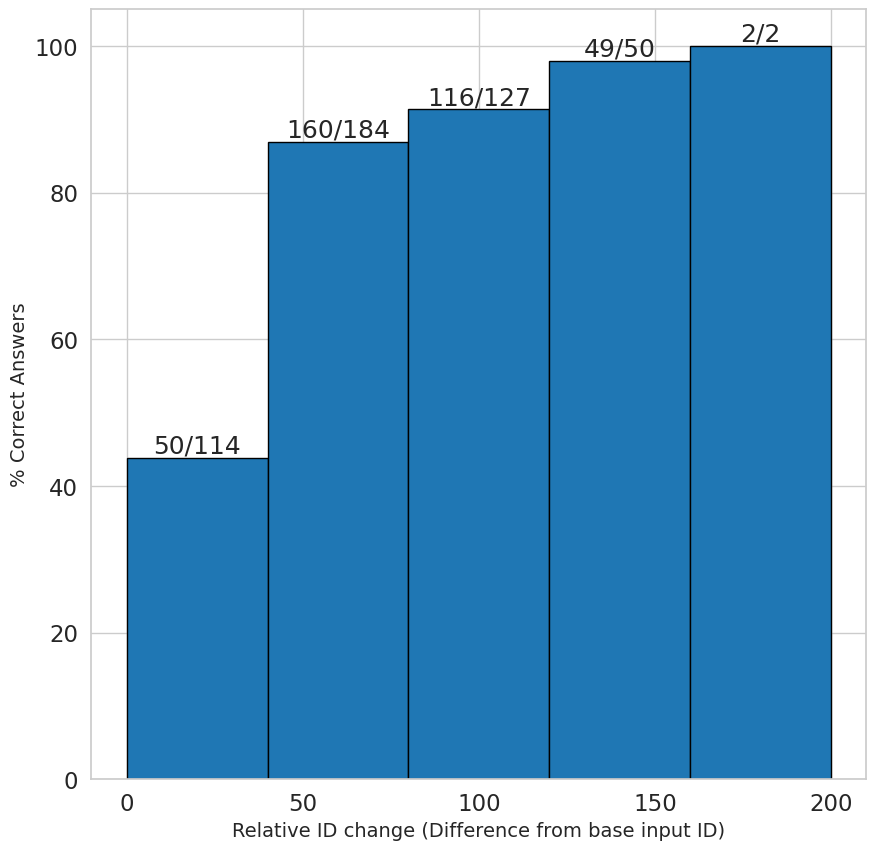}
    \end{subfigure}
    \caption{\textbf{Reasoning vs ID increase}. Percentage of correct responses, i.e., reasoning or extraction, concerning relative ID change for Llama3 8B (\textbf{Left}) and 70B (\textbf{Right}) Instruct models. The actual number of correct responses and the number of examples associated with each bin are denoted above each histogram for reference.
    We consider as input \emph{base prompt} examples with incorrect responses from the GSM8K-Zero dataset (approx. $300$ samples), along with their prepended variants where $1$ to $10$ fixed few-shot examples are used. For each input, we collect $(i)$ the change in the intrinsic dimension of the input concerning the base prompt, where the ID is computed at the final layer, and 
    $(ii)$ the correctness in the output generated by the LLM. We evaluate the response generated by prompting a Mixtral $8\times22$B Instruct model. We observe that the higher the ID change, the higher the probability of obtaining a correct response from the LLM.}
    \label{fig:llama3-fewshot}
\end{figure}

In this section, we are analyzing the capabilities of LLMs to answer reasoning questions through the lens of the aforementioned geometric analysis. Specifically, we are questioning how the increase in a number of regions induced by the MLP can lead to better reasoning capabilities. In fact, it is clear that approximation capabilities and generalization are not equivalent notions. However, it is not yet determined that the reasoning capabilities of LLMs are tied to their generalization. While these notions are still hard to pinpoint, we will focus in this experimental section on the relationship between intrinsic dimension, thus expressive power, and reasoning capabilities.

\begin{figure}[htbp]
    \begin{subfigure}{0.48\textwidth}
        \centering
                \textit{Llama3 8B}
        \vspace{0.2cm}
        \includegraphics[width=\textwidth]{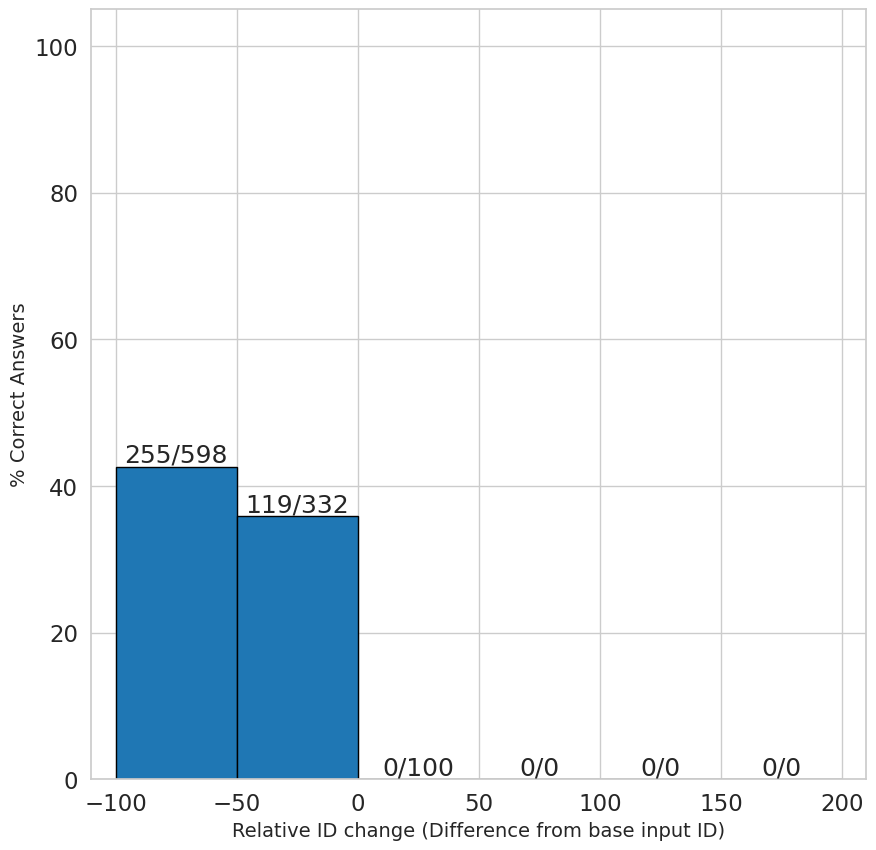}
    \end{subfigure}
    ~
    \begin{subfigure}{0.48\textwidth}
        \centering
                \textit{Llama3 8B}
        \vspace{0.2cm}
        \includegraphics[width=\textwidth]{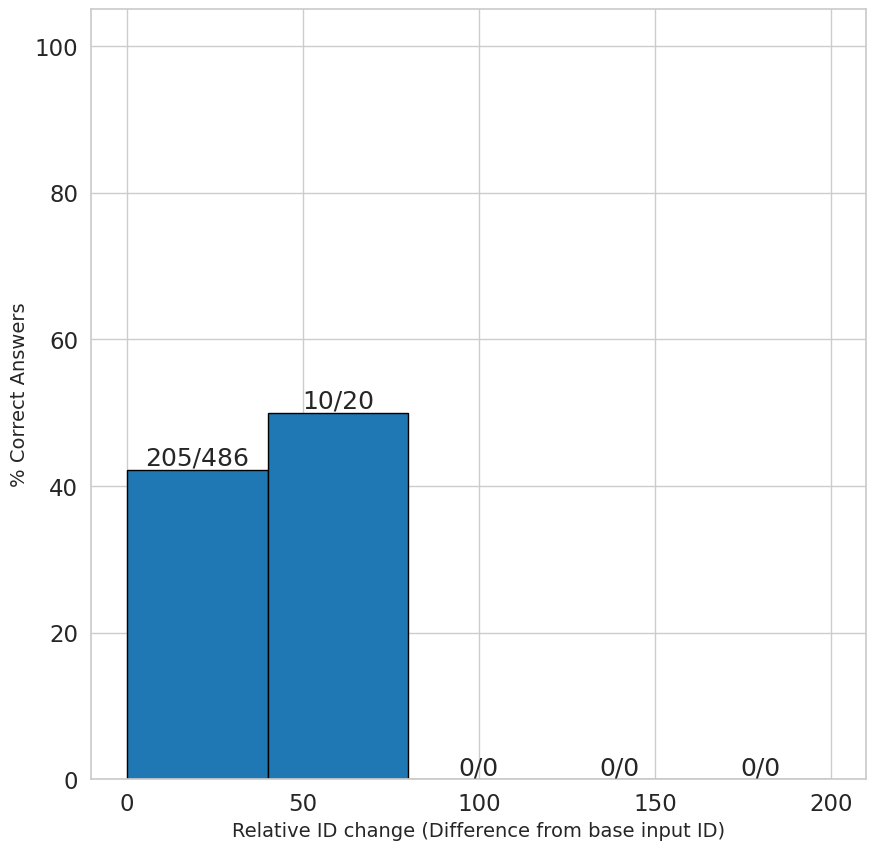}
    \end{subfigure}
    \caption{\textbf{Ablation with random tokens}. Percentage of correct responses, i.e., reasoning or extraction, concerning relative ID change for Llama3 8B Instruct model with random (\textbf{Left}) and shuffled few-shot example text (\textbf{Right}). 
    As in \autoref{fig:llama3-fewshot}, we consider as input \emph{base prompt} examples with incorrect responses from the GSM8K-Zero dataset (approx. $300$ samples), along with their prepended variants obtained through randomly sampled tokens or permuted text in the few-shot examples. We observe that the increase in ID is limited in the examples ($<60$) and even negative for the random token case. Consequently, obtaining a correct response is saturated and averages out to around 40\%, which is similar to the case with the 8B model and few-shot examples.}
    \label{fig:llama3-random}
\end{figure}

We propose two experiments to demonstrate that there is an intriguing correlation between them. For our experiments, we utilized the GSM8K-Zero dataset to assess the model's performance in generating correct answers across different few-shot scenarios, ranging from $0$ to $10$ shots. Specifically, for each sample and each $1$ to $10$-shot condition, we examined how the intrinsic dimension of the model varied across different layers when compared to the 0-shot baseline. Additionally, we evaluated how these variations influenced the quality of the model's responses. In the first experiment reported in \autoref{fig:llama3-fewshot}, the few shot examples are question-answer pairs randomly sampled from the GSM8K-Zero training set. For the second experiment reported in \autoref{fig:llama3-random}, these few shot examples are random tokens.

From these experiments, we make the following observations: $(i)$ pre-pending the question at hand with any type of token does increase the intrinsic dimension at the first layer. In fact, the first layer attention graph behaves as a uniform distribution over the tokens, however, this increase is not necessarily correlated with the reasoning capability of the model as the random token experiment demonstrates \autoref{fig:llama3-random}. $(ii)$ We observe that when the pre-pended tokens lead to an increase in the intrinsic dimension at the final layer of the model, the reasoning capabilities of the LLM improve significantly. This improvement is reflected in a higher percentage of questions being answered correctly.

\begin{figure}[htbp]
    \begin{subfigure}{0.48\textwidth}
        \centering
        \textit{Llama3 8B}
        \vspace{0.2cm}
        \includegraphics[width=\textwidth]{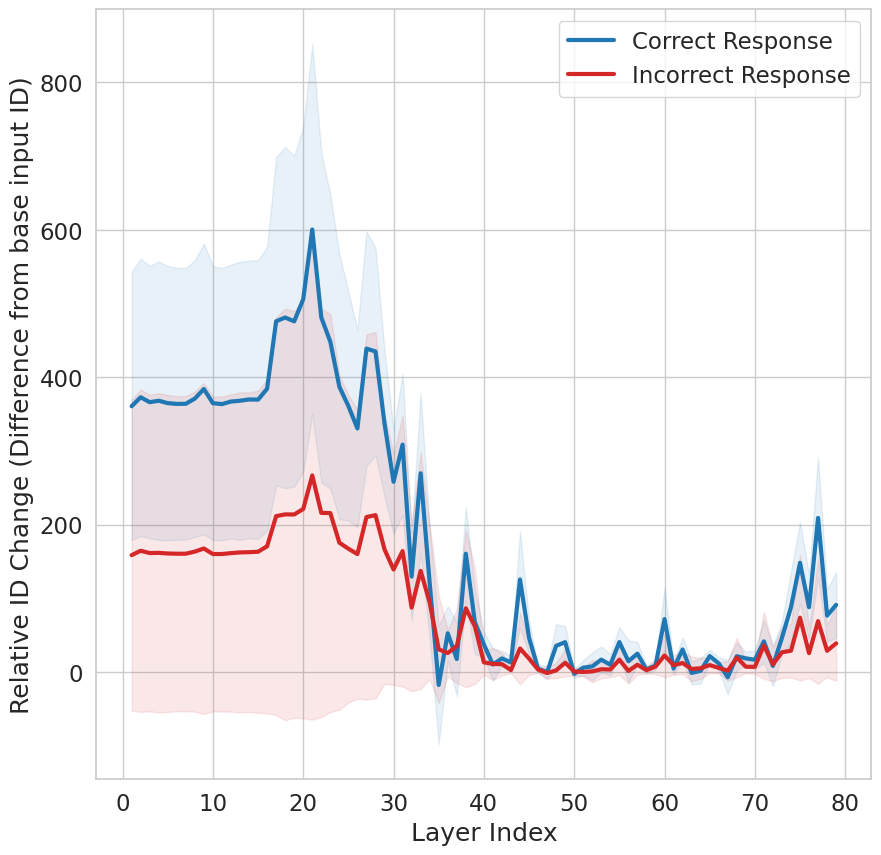}
    \end{subfigure}
    ~
    \begin{subfigure}{0.48\textwidth}
        \centering
        \textit{Llama3 70B}
        \vspace{0.2cm}
        \includegraphics[width=\textwidth]{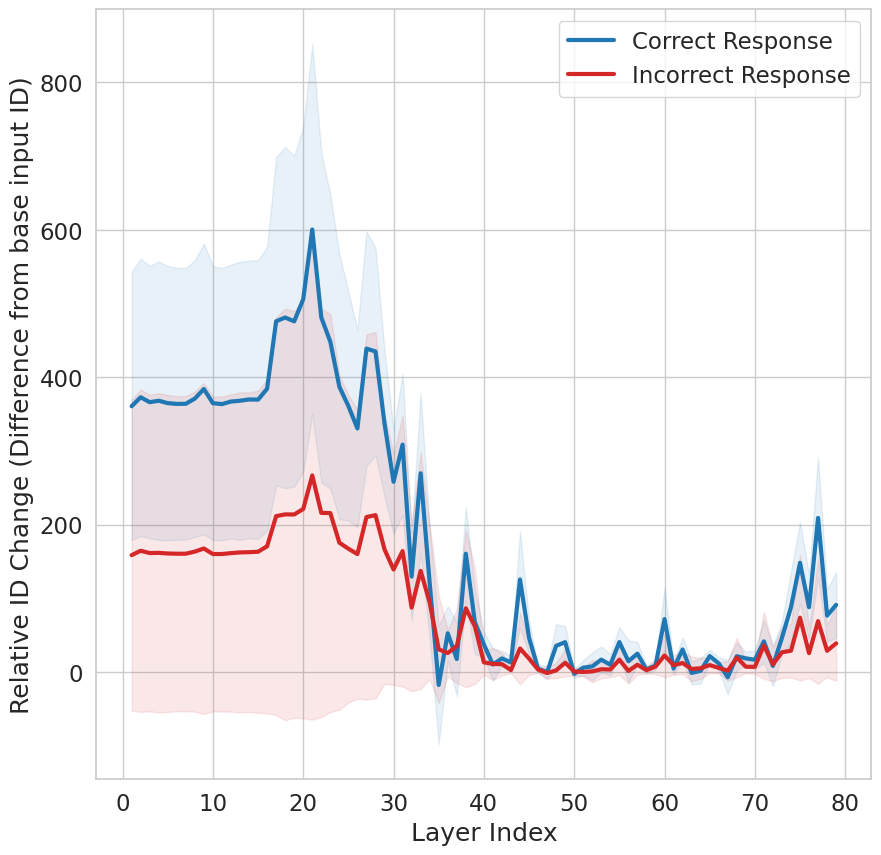}
    \end{subfigure}
    \caption{\textbf{Reasoning vs ID across layers}. Correct vs Incorrect response with respect to relative ID change for Llama3 8B (\textbf{Left}) and 70B (\textbf{Right}) Instruct models across each layer. 
    We consider as input \emph{base prompt} examples with incorrect responses from the GSM8K-Zero dataset (approx. $300$ samples), along with their prepended variants where $1$ to $10$ fixed few-shot examples are used. For each input, we collect $(i)$ the change in the intrinsic dimension of the input with respect to the base prompt, where the ID is computed at the final layer, and 
    $(ii)$ the correctness in the output generated by the LLM. We evaluate the response generated by prompting a Mixtral $8\times22$B Instruct model. We observe that the higher the ID change, the higher the probability of obtaining a correct response from the LLM.}
    \label{fig:llama3-across-layer}
\end{figure}

In \autoref{fig:llama3-across-layer}, we display the variation in intrinsic dimension of the $1$ to $10$ shots sampled with respect to $0$ for each layer. We clearly see that no matter the size of the model, the last layers ID are highly informative regarding the correctness of the response. While the first layers seem to have a huge variation in ID whether the output is correct or not, the variance is too large to be significant and reliable. 

These experiments highlight the correlation between a model's expressive power and its reasoning capabilities. As discussed in \autoref{sec:THEORY}, enhancing this expressive power can be achieved by increasing the dimension of the input to the MLP blocks. This relationship suggests that more complex input contributes to the improved reasoning performance of the model.

In LLMs, adding context to the prompt can increase the ID (depending on how related is the context to the question), and therefore increase the number of piece-wise affine maps produced by the MLP. One should note that, for an LLM, each token output by the self-attention head is independently transformed by the MLP. Thus, an MLP with a finer partition will have a more adaptive affine map for each token. If we think about this from an approximation standpoint, as the tokens are linearly combined to produce their predictions, the approximation error that is independently applied to each of them by the MLP can compound easily, and therefore, the more precise the partitioning around these tokens, the less the approximation error in the prediction. An aspect that has not been explored here as well as in most work is how these notions are tied to the generalization capabilities, if any, of LLMs.

In LLMs, incorporating additional context into the prompt can increase the intrinsic dimension of the model, particularly if the context is closely related to the question. This increase in ID leads to a greater number of piece-wise affine maps produced by the MLP. It’s important to note that in LLMs, each token output by the self-attention mechanism is independently transformed by the MLP. Consequently, an MLP with a more refined partitioning scheme will apply a more adaptive affine map to each token.

From an approximation perspective, since the model's predictions are formed by linearly combining these embedded tokens, the approximation error can accumulate across tokens. Therefore, finer partitioning around the tokens reduces the approximation error in the final prediction.

An intriguing aspect that remains largely unexplored in this work, as well as in most related research, is how these geometric insights into intrinsic dimension and affine map partitioning relate to the generalization capabilities of LLMs. This connection could offer valuable insights into the robustness and adaptability of these models in various contexts.

\section{Related Work}
The success of transformer-based models \cite{vaswani2017attention} across various input modalities has spurred significant research into the understanding of their internal mechanisms. Our work follows the lead of several key works on this topic. The difference, however, between these previous works and ours is the lens of analysis: we focus, fundamentally, on an end-to-end geometric perspective rather than a mechanistic framework \cite{elhage2021mathematical} or pattern analysis through empirical results \cite{voita2019analyzing, niu2021review, panigrahi2023task}. Our work is also different from these prior works in that we study the impact of model size and context length in transformer models and their role in reasoning capabilities, a critical aspect of modern LLMs whose understanding is largely absent.  

Theoretical works for understanding the reasoning capabilities of LLMs make use of input-output relationships through different frameworks in a domain-specific manner.
\cite{kim2022pure,sanford2024understanding, sanford2024transformers} make use of graph problems to understand the expressiveness of LLMs and associate them with the algorithmic complexity of the graph problem. \cite{zhou2022teaching, NEURIPS2023_6445dd88,liu2024exposing} use algorithmic reasoning as a way to understand the limitations of LLMs reasoning abilities. \cite{lee2023teaching}  investigate arithmetic learning and the impact of input formatting on LLM reasoning. Closely related, \cite{zhang2022unveiling} investigates the ability of LLMs to learn group actions. \cite{li2022systematic} consider a two-layer causal transformer and evaluate its generalization capability for copying, reversing, and sorting operations.

Other studies on transformers focus on initialization and training dynamics \cite{dong2021attention,  noci2022signal, boix2023transformers, trockman2023mimetic}. Albeit resorting to simplifying assumptions, these works shed light on the role of different components, such as the residual connection. 
The embedding geometry in the intermediate and last layers has also been explored previously. \cite{song2023uncovering} provides empirical insights about the position and context embeddings, \cite{song2023uncovering} presents an asymptotic (both in data and model) analysis to explain the emergent abilities of LLMs through latent space modeling, and \cite{hernandez2021low} identifies linear subspaces in contextualized embeddings to demonstrate geometric structure in LLMs.

Other works \cite{aghajanyan2020better, aghajanyan2020intrinsic, chen2020lottery} have studied the role of capacity in understanding LLMs and their transfer performance. In particular, \cite{aghajanyan2020intrinsic} empirically observed the role of intrinsic dimension (embedding dimension) in LLMs and its impact on generalization and downstream task representations. We note that our approach generalizes these observations while accommodating the sequence dimension, i.e., unlike previous works that relied on the dimension of entire sentences or tasks for their study, our geometric study presents a context-dependent analysis of LLMs.

Our work makes use of several mathematical tools developed with deep neural networks, in general, to understand transformer architecture. These observations, individually, may not be novel or have been implicitly noted in the literature. Notably, the spline view of neural networks was previously presented  \cite{balestriero2018spline}, which considered a partitioning of a \textit{fixed} dimensional input space by the non-linearities in the network.
Moreover, we note that the mathematical ideas presented in this work are likely implicitly known to researchers and practitioners familiar with transformers, and our contribution lies in leveraging this understanding to build a geometric interpretation of transformers.

\section{Discussion and Open Questions}
We presented here some aspects of DNNs and LLMs geometry, where in particular, we show the importance of the input space partitioning induced by the MLPs exploiting their piece-wise affine formulation. The adaptive partitioning of DNN in general plays a huge role in their approximation capability. In fact, as opposed to traditional spline, the regions induced by the MLP in their input space are data-dependent, and henceforth determined during training. We showed how such an interplay between approximation and the number of regions impacts the ability of LLMs to approximate functions. Then, we show that, while approximation power is not equivalent to generalization, it seems to be highly correlated to the reasoning capabilities of LLMs. In this work, we provided a brief overview of the underlying theory and a limited set of experiments related to these concepts. We believe that further exploration of this phenomenon is crucial to enhancing the reasoning capabilities of LLMs. Our hope is that through this, smaller LLMs can soon bridge the performance gap with their larger counterparts.


    

    




\bibliography{example_paper}
\bibliographystyle{ieeetr}

\end{document}